\documentclass{ifacconf}

\usepackage{balance}
\usepackage{graphicx}      
\usepackage{natbib}        
\usepackage{graphics} 
\usepackage{epsfig} 
\usepackage{amsmath} 
\usepackage{amssymb}  
\usepackage{color}
\usepackage{booktabs}
\usepackage{verbatim}
\usepackage{float}
\usepackage{enumitem}
\setlist{leftmargin=0.35cm}
\usepackage{fancyhdr}

\newtheorem{Thm}{\textit{Theorem}}
\newtheorem{df}{\textit{Definition}}

\newcommand{\pder}[2]{\ensuremath{\frac{\partial #1}{\partial #2}}}
\begin{document}
\begin{frontmatter}

\title{Fast, Composable Rescue Mission Planning for UAVs using Metric Temporal Logic} 

\thanks[footnoteinfo]{The research reported in this paper was partially supported by the US Office of Naval Research (ONR) Grant No. N00014-17-1-2622.}

\author[]{Usman A. Fiaz} \hspace{0.15 cm}
\author[]{John S. Baras} 

\address{Department of Electrical \& Computer Engineering, Institute for Systems Research, University of Maryland, College Park, MD, USA (e-mail: fiaz@umd.edu, baras@umd.edu)}

\begin{abstract}                
We present a hybrid compositional approach for real-time mission planning for multi-rotor unmanned aerial vehicles (UAVs) in a time critical search and rescue scenario. Starting with a known environment, we specify the mission using Metric Temporal Logic (MTL) and use a hybrid dynamical model to capture the various modes of UAV operation. We then divide the mission into several sub-tasks by exploiting the invariant nature of safety and timing constraints along the way, and the different modes (i.e., dynamics) of the UAV. For each sub-task, we translate the MTL specifications into linear constraints and solve the associated optimal control problem for desired path, using a Mixed Integer Linear Program (MILP) solver. The complete path for the mission is constructed recursively by composing the individual optimal sub-paths. We show by simulations that the resulting suboptimal trajectories satisfy the mission specifications, and the proposed approach leads to significant reduction in computational complexity of the problem, making it possible to implement in real-time. Our proposed method ensures the safety of UAVs at all times and guarantees finite time mission completion. It is also shown that our approach scales up nicely for a large number of UAVs.
\end{abstract}

\begin{keyword}
Flying robots, mission planning and decision making, safety, trajectory and path planning, optimal control of hybrid systems
\end{keyword}

\end{frontmatter}

\section{Introduction}

Multi-rotor Unmanned Aerial Vehicles (UAVs) in general and quadrotors in particular, find enormous applications in several key areas of research in academia as well as industry. These include but are not limited to search and rescue and disaster relief (\cite{waharte2010supporting, gholami2019drone}), autonomous aerial grasping and transport (\cite{hoareau2017package, fiaz2017passive}), and aerial coverage and surveillance (\cite{mozaffari2016efficient, fiaz2017thesis}). The motivation for this expanding zeal towards multi-rotor UAVs is twofold. First, these are highly inexpensive robots which can be extensively used as testbeds for much of the ongoing research in aerial robotics. Secondly, these are extremely agile robots, capable of much higher maneuverability in comparison with the other UAV classes, namely fixed-wing and helicopter style UAVs. This salient feature also edges them as a feasible platform to operate in congested environments, such as crowded city skies and constrained indoor workspaces.

The enormous impact of quadrotors in autonomous search and rescue becomes evident especially during natural disasters, where it is impossible for humans and terrestrial robots to access highly cluttered spaces to rescue people from life threatening situations. In such scenarios, these agile multi-rotor UAVs come to our rescue. They can either locate and grab a target themselves or can serve as a guide for other robots as well as humans to help evacuate a target. However, the need to evacuate the targets within limited, finite time cannot be more emphasized. Excess of the existing motion planning literature only guarantees successful collision avoidance for all times or successful and safe completion of the mission eventually i.e., without any guarantees on its finite time completion. However, time critical rescue missions require both safety and finite time completion guarantees. Although some recent works ensure finite time and safe mission completion for multi-agent systems, they are computationally complex and hence cannot be implemented in real-time. In addition they use simple dynamical models for the robots, which limit their maneuverability in practice.

In this paper, we propose a hybrid, compositional, optimization based method for real-time mission planning for quadrotors in a time critical search and rescue scenario. Starting with a known environment, we specify the mission using Metric Temporal Logic (MTL) and use a hybrid dynamical model to capture the various modes of UAV operation. We then divide the mission into several sub-tasks, by exploiting the invariant nature of safety and timing constraints along the way, and the different modes (i.e., dynamics) of the UAV. For each sub-task, we translate the MTL specifications into linear constraints, and solve the associated optimal control problem for desired path, using a Mixed Integer Linear Program (MILP) solver. The complete path for the mission is constructed recursively by composing the individual optimal sub-paths. We show by simulations that the resulting suboptimal trajectories satisfy the mission specifications, and the proposed approach leads to significant reduction in computational complexity of the problem, making it possible to implement in real time. 

Our proposed method ensures the safety of UAVs at all times and guarantees finite time mission completion. It is also shown that our approach scales up nicely for a large number of UAVs under some realistic assumptions on the environment.
Following are the main contributions of this paper:
\begin{itemize}
\item A hybrid optimization based framework for rescue mission planning for UAVs using MTL specifications under the assumption of known environment, and using a rich hybrid dynamical model for the UAVs.
\item Decomposition of complex MTL specifications into simpler MTL formulae and their translation into linear constraints.
\item Fast (i.e., real-time) and recursive computation of safe, composable, suboptimal trajectories for UAVs with finite time guarantees.
\item Limitations and scalability results for the proposed approach for large number of UAVs in a constrained environment.
\end{itemize}

The rest of the paper is organized as following. In Section 2, we provide a brief survey of the existing literature in relevant areas. Section 3 describes the essential notation and preliminaries on system dynamics, MTL specifications, and the workspace. In Section 4, we define the hybrid dynamical model for UAVs and describe some of its modes. In Section 5, we formulate the optimal control problem and detail our solution approach. Section 6 covers the results drawn from simulations. In Section 7, we summarize and analyze the outcomes of this work briefly, before concluding with some future prospects.

\section{Related Work}

Given any high level task, it is a standard practice in classic motion planning literature (\cite{latombe,lavalle}), to look for a set of trajectories, which the robot can follow, while satisfying the desired task specifications. This gives rise to the notion of optimal path planning, which considers an optimal path in the sense of optimizing some suitable cost function and finding a control law, to go from one position to another while satisfying some constraints (\cite{choset}). Traditionally, methods such as potential functions (\cite{xi}) have been used for multi-robot mission planning. However, they tend to fail in situations where the mission involves some finite time constraints or dynamic specifications. Aerial surveying of areas and time-critical search and rescue are two common examples of such tasks. 

Temporal logic (\cite{baier}) seems to address this problem, since it enables us to specify complex dynamic tasks in compact mathematical form. A bulk of modern motion planning literature is based on Linear Temporal Logic (LTL) (\cite{goerzen2010survey}), which is useful for specifying tasks such as visiting certain objectives periodically, surveying areas, ensuring stability and safety etc. (\cite{plaku2016motion, kantaros2018temporal}). However, from a control theory perspective, LTL only accounts for timing in the infinte horizon sense i.e., it can only guarantee something will \emph{eventually} happen and is not rich enough to describe finite time constraints. In addition, the traditional LTL formulation such as in (\cite{kress2009temporal}), assumes a static environment, which does not admit incorporating dynamic task specifications. 

On the other hand, Metric Temporal Logic (MTL) (\cite{MTL,MTL1}), can express finite time requirements between various events of the mission as well as on each event duration. This allows us to specify safety critical missions with dynamic task specifications and finite time constraints. An optimization based method for LTL was proposed in (\cite{KaramanCDC}), and (\cite{WolffICRA}), where they translate the LTL task specifications to Mixed Integer Linear Programming (MILP) constraints, which are then used to solve an optimal control problem for a linear point-robot model. This work was extended in (\cite{maity2015motion}), where the authors used bounded time temporal constraints using extended LTL for motion planning with linear system models. However, all these works did not incorporate a rich dynamical model of the robot, and also illustrated significant computational complexity issues for the proposed methods in case of planning for multiple robots in 3D. 

In some recent papers, optimization based methods with MTL specifications for single (\cite{zhou2015optimal}) and multiple (\cite{nikou2016cooperative}) robots, do guarantee safe and finite time mission completion. However, in both cases, the computation of the optimal trajectory is expensive (in the order of $\sim$500 sec computation time), and hence cannot be implemented in real-time. Moreover, these works put high constraints on the robot maneuverability, by limiting its dynamics to a simple linear (point-robot) model, which is in contradiction with the main reason for deploying quadrotors in constrained dynamic environments. Thus, in this paper, our intention is to use the rich dynamics of the robots in an intelligent way to divide and conquer the computationally complex problem of mission planning for multiple UAVs using finite time MTL constraints.

\section{Notation and Preliminaries}


\subsection{System Dynamics}
Any general (possibly nonlinear) dynamical system can be represented in the form:
\begin{align*}
	\dot{x}(t)=f(t,x,u)
\end{align*}
where for all time $t$ continuous
\begin{itemize}
\item $x(t) \in \mathcal{X} \subseteq \mathbb{R}^n$ 
	is the state vector of the system
\item $x_0 \triangleq x(0) \in \mathcal{X}_0 \subseteq \mathcal{X}$ 
	is the initial condition of the state vector and
\item $u(t) \in \mathcal{U} \subset \mathbb{R}^m$
	is the set of control inputs which is constrained in the control set $\mathcal{U}$.
\end{itemize}
Given a nonlinear model of the system, a linearization around an operating point $(x_*(t),u_*(t))$ is expressed as:
\begin{align*}
	\dot{\hat x}(t)=A(t)\hat x(t)+B(t)\hat u(t)
\end{align*}
where for all time $t$ continuous
\begin{itemize}
\item $\hat x(t) = x(t) - x_*(t)$
\item $\hat u(t) = u(t) - u_*(t)$
\item $A(t) = \pder{f}{x}(t)\Bigr|_{x=x_*,u=u_*}$
\item $B(t) = \pder{f}{u}(t)\Bigr|_{x=x_*,u=u_*}$
\end{itemize}
If time $t$ is discretized, then the system dynamics take the form:
\begin{equation} \label{eqn1}
x(t+1)=f(t,x(t),u(t))
\end{equation}
where as before, $x(t) \in \mathcal{X}$, $x(0) \in \mathcal{X}_0 \subseteq \mathcal{X}$, and $u(t) \in \mathcal{U}$ for all $t=0,1,2, \cdots$. 
Let us denote the trajectory for System~(\ref{eqn1}), with initial condition $x_0$ at $t_0$, and input $u(t)$ as: $\mathbf{x}_{t_0,x_0,u(t)}=\{x(t)~|~t\geq t_0,  x(t+1)=f(t,x(t),u(t)), ~ x(t_0)=x_0\}$. However, in this paper, for the sake of convenience, we use the shorthand notation i.e., $\mathbf{x}_{t_0}$ instead of $\mathbf{x}_{t_0,x_0,u(t)}$ to represent system trajectory whenever no explicit information about $u(t)$ and $x_0$ is required.
Similar to the the continuous time case, the corresponding linearized system in discrete time looks like:
\begin{align}
\label{eq:linearDyn}
	{\hat x}(t+1)=A(t)\hat x(t)+B(t)\hat u(t)
\end{align}
 for all $t=0,1,2, \cdots$. We use System~(\ref{eq:linearDyn}) form dynamics in our problem formulation in Section 5.

\subsection{Metric Temporal Logic (MTL)}
The convention on MTL syntax and semantics followed in this paper is the same as presented in (\cite{MTL}). More details on specifying tasks as MTL formulae can also be found in (\cite{MTL1}).
\begin{df}\label{defat}
\textit{An atomic proposition is a statement with regard to the state variables $x$ that is either $\mathbf{True}~(\top)$ or $\mathbf{False}~( \perp)$ for some given values of $x$.} 
\end{df}
Let $\Pi =\{ \pi_1, \pi_2, \cdots \pi_n \}$ be 
the set of atomic propositions which labels $\mathcal{X}$ as a collection of areas of interest in some workspace, which can possibly be time varying. Then, we can define a map $L$ which labels this workspace or environment as follows:
\begin{equation} \label{fdef} \notag
L: \mathcal{X} \times \mathcal{I} \rightarrow 2^{\Pi}
 \end{equation}
where $\mathcal{I} =\{[t_1,t_2]~|~t_2  > t_1 \geq 0 \} $ and $2^{\Pi}$ denotes the power set of $\Pi$ as usual. In general, $\mathcal{I}$ represents an interval of time but it may just also represent a time instance. 
For each trajectory of System~(\ref{eq:linearDyn}) i.e., $\mathbf{x}_{t_0}$ as before, the corresponding sequence of atomic propositions, which $\mathbf{x}_{t_0}$ satisfies is given as: $\mathcal{L}(\mathbf{x}_{0})=L(x(0),0)L(x(1),1)...$.

We later specify the tasks formally using MTL formulae, which can incorporate finite timing constraints. These formulae are built on the stated atomic propositions (Definition~\ref{defat}) by following some grammar.
\begin{df} \label{def1}
 \textit{The syntax of MTL formulas are defined in accordance with the following rules of grammar:}
 \begin{center}
 $\phi ::= \top ~| ~\pi~ |~\neg \phi~ | ~\phi \vee \phi ~|~\phi \mathbf{U}_I \phi  ~ $
 \end{center} 
 \end{df}
where $I\subseteq [0, \infty]$, $\pi \in \Pi$, $\top$ and $\neg\top (=\bot)$ are the Boolean constants for $true$ and $false$ respectively. $\vee$ represents the disjunction while $\neg$ represents the negation operator. $\mathbf{U}_I$ denotes the Until operator over the time interval $I$. Similarly, other operators (both Boolean and temporal) can be expressed using the grammar imposed in Definition \ref{def1}. Some examples are conjunction ($\wedge$), always on $I$ ($\Box_I$), eventually within $I$ ($\Diamond_I$) etc. Further examples of temporal operators can be found in (\cite{KaramanCDC}).

\begin{df}\label{mtlsym}
 \textit{The semantics of any MTL formula $\phi$ is recursively defined over a trajectory $x_t$ as:\\
 $x_{t} \models \pi$ iff $\pi \in L(x(t),t)$\\
 $x_{t} \models \neg \pi$ iff $\pi \notin L(x(t),t)$\\
 $x_{t} \models \phi_1\vee \phi_2$ iff $x_{t} \models \phi_1$ or $x_{t} \models \phi_2$\\
 $x_{t} \models \phi_1\wedge \phi_2$ iff $x_{t} \models \phi_1$ and $x_{t} \models \phi_2$\\
 $x_{t} \models \bigcirc \phi$ iff $x_{t+1} \models \phi$\\
 $x_{t} \models \phi_1\mathbf{U}_I \phi_2$ iff $\exists t' \in I$ s.t. $x_{t+t'} \models  \phi_2$ and $\forall$ $t'' \leq t'$,\\ $ x_{t+t''} \models \phi_1$}.
\end{df}
Thus, for instance, the expression $\phi_1 \mathbf{U}_I \phi_2$ means the following: $\phi_2$ is true within time interval $I$, and until $\phi_2$ is true, $\phi_1$ must be true. Similarly, the MTL operator $\bigcirc \phi$ means that $\phi$ is true at next time instance.  $\Box_I \phi$  means that $\phi$ is always true for the time duration or during the interval $I$, $\Diamond_I \phi$ implies that $\phi$ eventually becomes true within the interval $I$. More complicated formulas can be specified using a variety of compositions of two or more MTL operators. For example, $\Diamond_{I_1} \Box_{I_2} \phi$ suffices to the following: within time interval $I_1$, $\phi$ will be eventually true and from that time instance, it will always be true for an interval or duration of $I_2$. The remaining Boolean operators such as implication ($\Rightarrow$) and equivalence ($\Leftrightarrow$) can also be represented using the grammar rules and semantics given in Definition~\ref{def1} and Definition~\ref{mtlsym}. 
Similar to the convention used in Definition~\ref{mtlsym}, a system trajectory $\mathbf{x}_{t_0}$ satisfying an MTL specification $\phi$ is denoted as $\mathbf{x}_{t_0} \models \phi$. 

\begin{figure}
	\centering
		\includegraphics[width=8.6 cm]{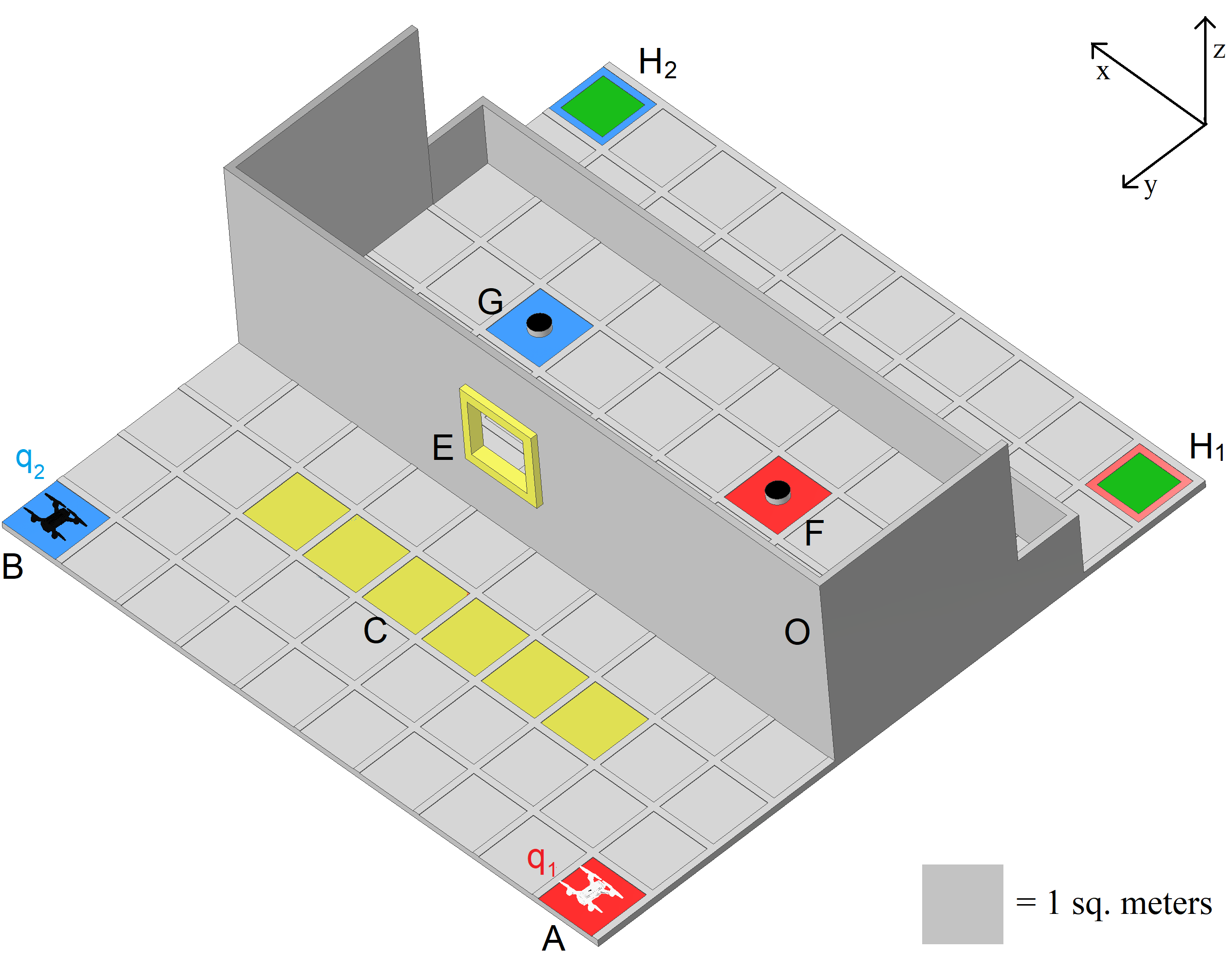}
		\caption{CAD model for the workspace used. The environment is a 10x10x3 $m^3$ workspace which is divided into several 2D regions of interest that are labeled with alphabets and marked with different colors.}
	\label{fig:workspace}
\end{figure}

\subsection{The Workspace}

Throughout this paper, we repeatedly refer to a time critical rescue mission defined on the constrained workspace shown in Fig.~\ref{fig:workspace}. It is a custom built CAD environment designed with the intention to closely suit our problem. As shown in Fig.~\ref{fig:workspace}, various areas of interest are marked on the workspace using different alphabets. The role of each of these areas of interest will become obvious as we formulate the problem and specify the mission using MTL specifications in Section 5. For now, we briefly describe the mission in a two UAV setting. Starting from initial positions $A$ and $B$, two quadrotor UAVs $q_1$ and $q_2$ need to rescue two objects located at $F$ and $G$ respectively from a constrained environment. The objects are accessible only through a window $E$, with dimensions such that it allows only one UAV to pass at a given time. Therefore, one of the UAVs has to wait at $C$ for the other to pass first. It is assumed that there are no additional obstacles in the area other than the walls $O$ and the UAVs themselves. The task for each UAV is to grasp its respective target object, and transport it to safety (marked $H_1$ and $H_2$ respectively for $q_1$ and $q_2$) in given finite time. While doing so, the UAVs need to avoid the obstacles $O$ as well as each other, in particular at the window $E$. Later on, we also use prime region notation; e.g., $A'$ to represent the same 2D region $A$. There $A'$ represents an altitude (w.r.t. $z$-axis) variation of the quadrotor while it is in the same 2D region $A$.

\section{Quadrotor Dynamics}

We adopt the generalized nonlinear model for the quadrotor presented in (\cite{Kumar}). We build a hybrid model for the system with five linear modes, namely \emph{Take off}, \emph{Land}, \emph{Hover}, \emph{Steer}, and a task specific \emph{Grasp} mode. The linearization for each mode is carried out separately about a different operating point. This enables our system to have rich dynamics with less maneuverability restrictions, while each mode still being linear. This is an important point and its significance becomes apparent once we formulate the problem and present our solution approach, since it requires all constraints in the problem to be linear. (see Section 5). 

\subsection{General Nonlinear Model}
The dynamics of a quadrotor can be fully specified using two coordinate frames. One is a fixed earth (or world) frame, and the second is a moving body frame. Let the homogeneous transformation matrix from body frame to earth frame be $R(t)$, which is a function of time $t$. In state space representation, the quadrotor dynamics are represented as twelve states namely $[x,y,z,v_x,v_y,v_z,\phi,\theta,\psi,\omega_{\phi},\omega_{\theta},\omega_{\psi}]^T$, where $\xi=[x,y,z]^T$ and $v=[v_x,v_y,v_z]^T$ represent the position and velocity of the quadrotor respectively with respect to the body frame. $[\phi,\theta,\psi]^T$ are the angles along the three axes (i.e., roll, pitch, and yaw respectively), and $\Omega=[\omega_{\phi},\omega_{\theta},\omega_{\psi}]^T$ represents the vector containing their respective angular velocities. Under the rigid body assumptions on its airframe, the Newton-Euler formalism for quadrotor in earth frame is given by: 
\begin{align} \label{quadrotor}
 \dot{\xi} & =v  \nonumber \\
 \dot{v} & = -g\mathbf{e_3} + \frac{F}{m}R\mathbf{e_3} \\
 \dot R &=R \hat{\Omega} \nonumber \\
 \dot\Omega &= J^{-1} (-\Omega \times J\Omega + \tau)\nonumber
\end{align}
where $J$ is the moment of inertia matrix for the quadrotor, $g$ is the gravitational acceleration, $\mathbf{e_3}=[0, 0, 1]^T$, $F$ is the total thrust produced by the four rotors, and $\tau=[\tau_x,\tau_y,\tau_z]^T$ is the torque vector, whose components are the torques applied about the three axes. So, $F$, $\tau_x$, $\tau_y$, and $\tau_z$ are the four control inputs to System~(\ref{quadrotor}).

\subsection{Hybrid Model with Linear Modes}

\begin{figure}[]
	\centering
		\includegraphics[width=8.6 cm]{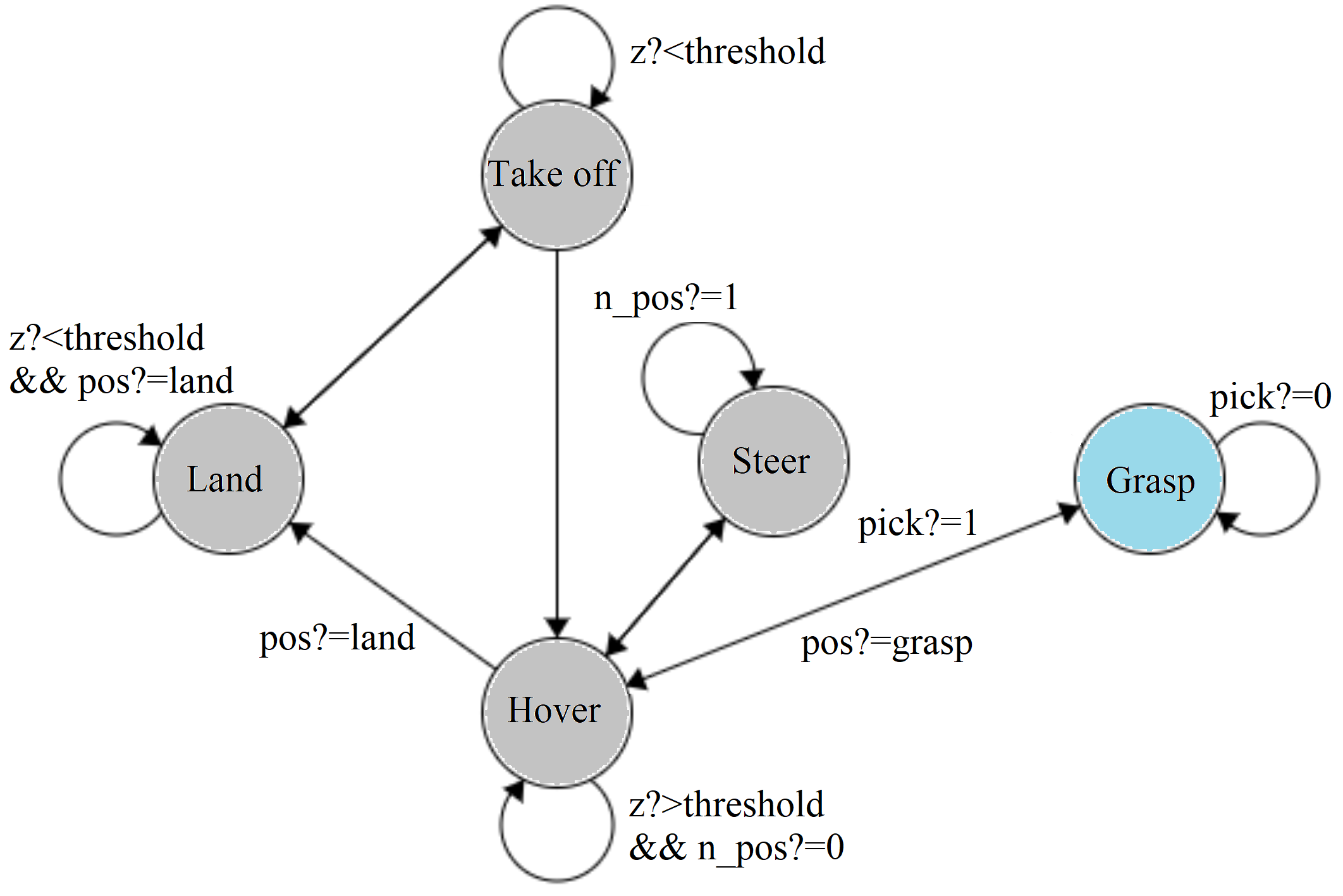}
		\caption{The simplified hybrid dynamical model for the quadrotor. Some guard conditions are hidden for readability. We use linearized dynamics around different operating points for each mode. This makes the model rich in dynamics as well as linear at the same time.}
	\label{fig:hybrid}
\end{figure}

System~(\ref{quadrotor}) serves as a starting point for generating a hybrid model for the quadrotor with five modes, which are represented by a labeled transition system as shown in Fig.~\ref{fig:hybrid}. As usual, the states (or modes) denote the action of the UAV, such as \emph{Take off} and \emph{Steer}, while the edges represent the change or switch to another action. The change is governed by some suitable guard condition. Note, that some edges donot exist; for example, the quadrotor cannot go from \emph{Land} to \emph{Hover} without taking the action \emph{Take off}. Each state of the transition system follows certain dynamics, which result from a linearization of System~(\ref{quadrotor}) around a different operating point. For example, consider the \emph{Hover} mode. One possible choice of operating points for the linearization of system dynamics in this mode is $\psi = 0$. This implies that the two states $\psi$ and $\omega_{\psi}$ i.e., the yaw angle and its respective angular velocity are identically zero, and thus can be removed from the state space representation. Consequently, the state space dimension is reduced to ten, and the control set is reduced to three inputs as well; i.e., $F, \tau_x$, and $\tau_y$. The resulting linearized model can be written in standard (discrete time) form as: ${\sigma(t+1)} = A(t)\sigma(t) + B(t)\gamma(t)$, where $\sigma(t)$ is the state, and $\gamma(t)$ is the input (in vector notation), with the two system matrices given as:
\vspace{0.3cm}\\
 $\hspace*{0.5cm} A=\begin{bmatrix} \mathbf{0} & I & \mathbf{0} & \mathbf{0} \\ 
 \mathbf{0} & \mathbf{0} & \begin{bmatrix} 0  & g \\ -g & 0\\0 & 0 \end{bmatrix} & \mathbf{0}\\ 
 \mathbf{0} &\mathbf{0} &\mathbf{0} & I \\ \mathbf{0} &\mathbf{0} &\mathbf{0} &\mathbf{0}\\
 \end{bmatrix}; \hspace{0.25cm} B = \begin{bmatrix} \mathbf{0} & \mathbf{0}\\ \begin{bmatrix} 0\\  0 \\ 1/m \end{bmatrix} &\mathbf{0} \\ \mathbf{0} & \mathbf{0} \\\mathbf{0} & I_{2 \times 3} J^{-1}\\ \end{bmatrix}$
\vspace{0.3cm}\\
where $I_{2,3}= [I_{2,2}~~\mathbf{0}_{2,1}]$, and all zero and identity matrices in $A(t)$ and $B(t)$ are of proper dimensions. We adopt similar procedure to linearize System~(\ref{quadrotor}) around other operating points for different modes, and obtain linearized dynamics for the hybrid model. Here, we omit the discussion about the selection of these operating points, but it can be found in (\cite{Garcia}).

\begin{figure}[]
	\centering
		\includegraphics[width=8.0 cm]{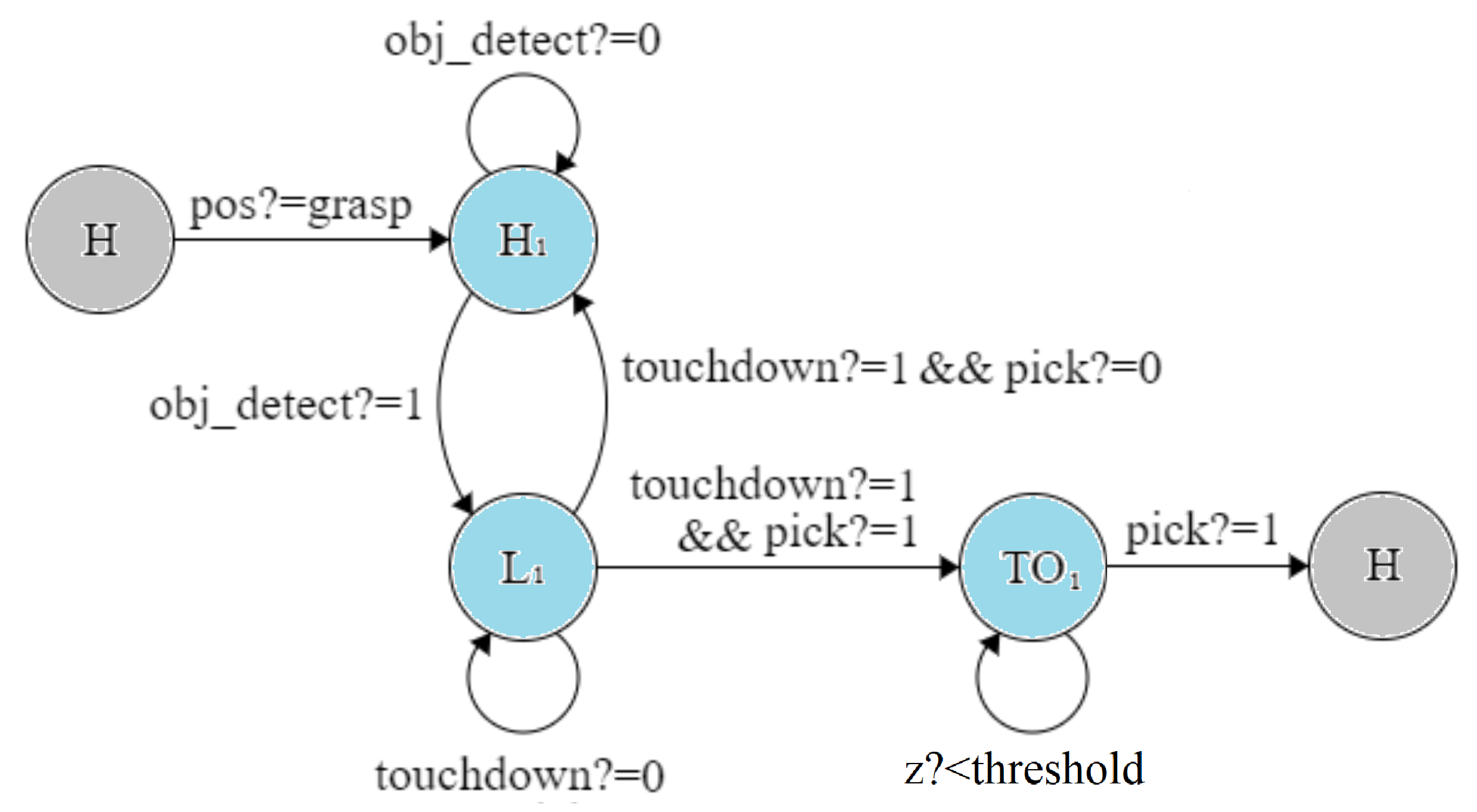}
		\caption{The \emph{Grasp} mode expressed as a combination of \emph{Hover} ($H_1$), \emph{Land} ($L_1$), and \emph{Take off} ($TO_1$) modes (colored cyan), with special guard conditions.}
	\label{fig:grasp}
\end{figure}

\subsection{The Grasp Mode}
Grasping in general is a very challenging problem, in particular when dexterity based manipulation is involved. Since \emph{Grasp} is the only task specific dynamical mode of our system, it was advisable to simplify the grasping routine within the high level task. However, in case of aerial grasping, some passive mechanisms (\cite{fiaz2019impulsive}) have been shown to be very reliable in grasping an object with an instantaneous touchdown onto its surface (\cite{fiaz2018intelligent}). Thus, under this reliable passive aerial grasping assumption (i.e., instantaneous touchdown and grasp), we can express the \emph{Grasp} mode as a switching combination of \emph{Hover}, \emph{Land}, and \emph{Take off} dynamics with special guard conditions. This clearly simplifies the problem of having the need to introduce a complex gripper and its end-effector dynamics into the \emph{Grasp} mode of the hybrid model. Figure~\ref{fig:grasp} depicts this representation of the \emph{Grasp} mode in terms of the \emph{Hover}, \emph{Land}, and \emph{Take off} dynamics.

\section{Method: Formulation and Solution}
Given the map of the environment, we can write down a mission specification for each quadrotor as an MTL formula $\phi_i$. For the workspace described in Section 3, a possible MTL task specification for the $i^{th}$ quadrotor $q_i$ can be written as: \\
\vspace{-0.2cm}
\[\phi_i = \Diamond_{[0,T_1]}(Object~Location) \wedge \Box_{[0,T_2]}(Object~Location)' \] \\ 
\vspace{-0.75 cm} 
\hspace*{0.5cm}\[ \wedge \Diamond_{[0,T_3]}\Box (Safe~Location) \wedge \Box \neg (Obstacles) \wedge \Box\neg q_j
\] 

where $i,j \in \{1, 2, . . . , N\}$, $i \neq j$, and $q_j \in \cal{N}_i(t)$. $\cal{N}_i(t)$ represents the neighborhood set of quadrotor $q_i$ i.e., all quadrotors $q_j$ s.t. $||q_i - q_j|| \leq \rho$, for some $\rho > 0$. $T_3$, $T_2$, $T_1$ are discrete time units, and the prime notation models the grasping action within the same area of the workspace with some altitude variation of the UAV along the $z$-axis. Thus, for instance, in the two UAV setting as shown in Fig.~\ref{fig:workspace}, a possible MTL specification for quadrotor $q_1$ is:\\
\vspace{-0.2cm}
\[\phi_1 = \Diamond_{[0,T_1]}(F) \wedge \Box_{[0,T_2]}(F)'  \wedge \Diamond_{[0,T_3]}\Box (H_1) \wedge \Box \neg (O) \wedge \Box\neg q_2
\] 

A similar mission specification $\phi_2$ can be written for the quadrotor $q_2$ and so on. Now using the quadrotor dynamics (\ref{eq:linearDyn}) and these MTL formulae $\phi_i$, we can state the rescue mission planning as an optimal control problem.

\subsection{Problem Statement and Formulation}
We set up the described mission as a standard optimal control problem in discrete time. Given the system dynamics (\ref{eq:linearDyn}), the objective is to find a suitable control law that steers each quadrotor through some regions of interest in the workspace within desired time bounds, so that it evacuates the target safely to a desired location. This control also optimizes some cost function, while the associated task constraints are specified by an MTL expression. As before, let $\phi_i$ denote the MTL formula for the mission specification, and $J(x_i(t,u_i(t)),u_i(t))$ be the cost function to be minimized. Then, the corresponding optimization problem for quadrotor $q_i$, $i \in \{1, 2, . . . , N\}$ is given by:
\begin{prob}
\[\begin{array}{c c} \underset{x_i,u_i}{\min} \text{ }   &~J(x_i(t,u_i(t)),u_i(t)) \\ \text{s.t.  } & x_i(t+1)=A(t)x_i(t)+B(t)u_i(t) \\ & \mathbf{x_i}_{t_0} \models \phi_i\\ \end{array}\]
\end{prob}
Problem 1 is a discrete time optimal control problem with linear dynamics. However, it includes a complex MTL satisfiability constraint. In addition, notice that in our case, the hybrid model of the quadrotor has multiple linear modes. Therefore, in its current form, Problem 1 is not directly solvable. 

We now describe two methods that transform Problem 1 into a set of readily solvable Mixed Integer Linear Programs (MILPs). First, we describe a method for translating the MTL satisfiability constaint into a set of linear constraints. This approach renders Problem 1 solvable as a MILP for a linear cost and a given dynamical mode of the system. Secondly, we decompose the complex MTL specification $\phi_i$ into a set of simpler MTL formulae $\phi_i^k$, $i \in \{1, 2, ..., N\}$ and $k \in \{1, 2, ..., M\}$. This suffices to breaking down of the original problem into $M$ sub-problems, each with a set of linear constraints and exactly one associated dynamical mode of the hybrid system. Combining both these methods, the resulting $M$ MILPs are expected to have significantly reduced computational complexity than the parent problem.

\subsection{MTL Formulae to Linear Constraints}
This method is based on the approach presented in (\cite{KaramanCDC}) where the authors translate LTL specifications into linear constraints. Along similar lines, we now present an approach to translate MTL specifications into mixed integer linear constraints. We start with a simple temporal specification and work through the procedure to convert it to mixed integer linear constraints. We then use this example as a foundation for translating the MTL operators into equivalent linear constraints.

Consider the constraint that a trajectory $x(t)$ lies within a convex polytope $\mathcal{K}$ at time $t$. Since $\mathcal{K}$ is convex, it can be represented as an intersection of a finite number of halfspaces. A halfspace can be represented as set of points, $\mathcal{H}_i=\{ x: ~ h_i^Tx\leq a_i\}$. Thus, $x(t) \in \mathcal{K}$ is equivalent to $x(t) \in \cap_{i=1}^n \mathcal{H}_i =\cap_{i=1}^n \{ x: ~ h_i^Tx \leq a_i \}$. So, the constraint  $x(t) \in \mathcal{K}$ $\forall$ $t \in \{t_1, t_1+1, \cdots t_1+n \}$ can be represented by the set of linear constraints $\{ h_i^Tx(t)\leq a_i \}$, $\forall$ $ i =\{1,2,\cdots, n\}$ and $\forall t\in \{t_1, t_1+1, \cdots t_1+n \}$.
 
In a polytopic environment, atomic propositions (see Definition~\ref{def1}), $p,q \in \Pi$, are related to system state via conjunction and disjunction of linear halfspaces (\cite{KaramanCDC}). Let us consider the case of a convex polytope and let $b_i^t \in \{0,1\}$ be some binary variables associated with the corresponding halfspaces $\{x(t) : h_i^T x(t)\leq a_i\}$ at time $t=0,...,N$. We can then force the constraint: $b_i^t=1$ $\iff$ $h_i^T x(t) \leq a_i$, by introducing the following linear constraints:
\begin{equation} 
 h_i^T x(t) \leq a_i + M(1-b_i^t) 
\end{equation} 
\[ h_i^T x(t) \geq a_i -M b_i^t +\epsilon\]
where $M$ and $\epsilon$ are some large and small positive numbers respectively. If we denote $K_t^\mathcal{K}=\wedge_{i=1}^n b_i^t$, then $K_t^{\mathcal{K}}=1$ $\iff$ $x(t) \in \mathcal{K}$. This approach is extended to the general nonconvex case by convex decomposition of the polytope. Then, the decomposed convex polytopes are related using disjunction operators. Similar to conjunction, as is described later in this section, the disjunction operator can also be translated to mixed integer linear constraints.

Let $S_\phi(x, b, u, t)$ denote the set of all mixed integer linear constraints corresponding to a temporal expression $\phi$. Using the described procedure, once we have obtained $S_p(x,b,u,t)$ for atomic propositions $p \in \Pi$, we can formulate $S_\phi (x,b,u,t)$ for any MTL formula $\phi$. 
Now, for the Boolean MTL operators, such as $\neg$, $\wedge$, $\vee$, let $t \in \{0,1,...,N\}$, and as before $K_t^\phi \in [0,1]$ be the continuous variables associated with the formula $\phi$ generated at time $t$ with atomic propositions $p \in \Pi$. Then 
$\phi=\neg p$ is the negation of an atomic proposition, and it can be modeled as:
	\begin{align} \label{nega} 
	K_t^{\phi} = 1- K_t^p
	\end{align}
 the conjunction operator, $\phi = \wedge^m_{i=1}p_i$, is modeled as:
	\begin{align}
	K_t^{\phi} & \leq K_t^{p_i}, \quad i=1,...,m \\ \nonumber 
	K_t^{\phi} & \geq 1-m+ \sum_{i=1}^m {K_t^{p_i}} 
	\end{align}
and the disjunction operator, $\phi = \vee^m_{i=1}p_i$, is modeled as:
	\begin{align}
	K_t^{\phi} & \geq K_t^{p_i}, \quad i=1,...,m \\ \nonumber 
	K_t^{\phi} & \leq \sum_{i=1}^m {K_t^{p_i}}
	\end{align}
Similar to binary operators, temporal operators such as $\Diamond, \Box$, and $\mathcal{U}$ can be modeled using linear constraints as well. Let $t \in \{0,1,...,N-t_2\}$, where $[t_1,t_2]$ is the time interval used in the MTL specification $\phi$. Then, 
eventually operator: $\phi=\Diamond_{[t_1,t_2]} p$ is modeled as:
	\begin{align}
	K_t^{\phi} & \geq K_\tau^{p}, \quad \tau \in \{t+t_1,...,t+t_2 \} \\ \nonumber
	K_t^{\phi} & \leq \sum_{\tau=t+t_1}^{t+t_2} {K_\tau^{p}} 
	\end{align}
and always operator: $\phi=\Box_{[t_1,t_2]} p$ is represented as:
	\begin{align}\label{always}
	K_t^{\phi} & \leq K_\tau^{p}, \quad \tau \in\{t+t_1,...,t+t_2\}\\ \nonumber
	K_t^{\phi} & \geq \sum_{\tau=t+t_1}^{t+t_2} {K_\tau^{p}} -(t_2-t_1)
	\end{align}
	
and until operator: $\phi=p~\mathcal{U}_{[t_1,t_2]}~q$ is equivalent to:
	\begin{align}\label{eq:until}
	c_{tj} & \leq K_q^{j} \quad j \in \{t+t_1,\cdots,t+ t_2\}\nonumber \\ \nonumber
	c_{tj} & \leq K_p^{l} \quad l \in \{t, \cdots, j-1\}, j \in \{t+t_1,\cdots,t+ t_2\}\\ \nonumber
	c_{tj} & \geq K_q^{j} + \sum_{l=t}^{j-1} K_p^l -(j-t) \quad j \in \{t+t_1,\cdots,t+ t_2\}\\
	c_{tt} & = K_q^t  \\ \nonumber
	K_t^{\phi} & \leq \sum_{j=t+t_1}^{t+t_2}c_{tj} \\ \nonumber
	K_t^{\phi} & \geq c_{tj} \quad j \in \{t+t_1,\cdots, t+t_2\} \nonumber
	\end{align}
The equivalent linear constraints for until operator (\ref{eq:until}) are constructed using a procedure similar to (\cite{KaramanCDC}). The modification for MTL comes from the following result in (\cite{MTL}).
\[K_t^{\phi}= \bigvee_{j=t+t_1}^{t+t_2}\left((\wedge_{l=t}^{l=j-1}K_p^l)  \wedge K_q^j \right). \]  

All other combinations of MTL operators for example, eventually-always operator: $\phi=\Diamond_{[t_1,t_2]} \Box_{[t_3,t_4]}p$ and always-eventually operator: $\phi=\Box_{[t_1,t_2]}\Diamond_{[t_3,t_4]} p$ etc., can be translated to similar linear constraints using (\ref{nega})-(\ref{eq:until}). In addition to the collective operator constraints, we need another constraint $K_0^\phi=1$ as well, which suffices to the overall satisfaction of a task specification $\phi$. 

Using this approach, we can translate an MTL formula $\phi$ into a set of mixed integer linear constraints $S_\phi (x,b,u,t)$, which converts the associated optimal control problem (e.g. Problem 1) to a MILP for some linear cost function.

\subsection{Decomposition of Complex MTL Formulae}
Notice that the worst case complexity of the above MILP (Problem 1 with linear constraints and linear cost) is exponential i.e., $O(2^{mT})$, where $m$ is the number of boolean variables or equivalently the number of halfspaces required to express the MTL formula, and $T$ is the discrete time horizon. Therefore, it is logical to consider decomposing the task specification $\phi_i$ into several simpler sub-tasks $\phi_i^k$, where $i \in \{1, 2, ..., N\}$ and $k \in \{1, 2, ..., M\}$.

In (\cite{schillinger2018decomposition}), the authors proposed a timed-automata based approach to decompose a complex LTL specification into a finite number of simpler LTL specifications. In case of MTL, this method is not applicable in general, because MTL to Buchi automata conversion is not always possible. However, in the special case, where the finite time constraints are specified as intervals i.e., in case of Metric Interval Temporal Logic (MITL), we can break down a complex MTL specification (mission) into finite number of simpler MTL specifications (sub-tasks), if the sum of the finite timing intervals in decomposed MTL specifications does not violate the finite timing interval of the original MTL specification. We can state this proposition as the following theorem:

\begin{Thm}
\textit{Given an MITL specification $\phi_i$, there exists some finite $M$-length decomposition $\phi_i^k$, $k \in \{1, 2, ..., M\}$, s.t. $\wedge_{k=1}^{M}(\phi_i^k) \implies \phi_{i}$, if $\sum_{k=1}^M T_k \leq T_i$, where $T_i$ is the finite timing interval for $\phi_{i}$, and $T_k$s are the corresponding finite timing intervals for $\phi_i^k$, $ \forall k \in \{1, 2, ..., M\}$.}
\end{Thm}

The proof follows directly from Theorem 2 in (\cite{schillinger2018decomposition}). It can also be readily verified using a timed automata simulation in UPAAL (\cite{behrmann2006uppaal}). 

Notice that Theorem 1 does not guarantee an equivalence relationship between $\phi_i$ and $\phi_i^k$. Moreover, this decomposition is not unique. However, if the mission is specified by an MITL specification and the system is represented by a hybrid model, which is the case in our problem, one convenient design choice is to pick such $\phi_i^k$, for which there is only one associated dynamical mode of the system. We provide examples of such decomposition in Section 6.

For example, consider again the workspace in Fig.~\ref{fig:workspace} and let $\phi_1$ be given by: 
\[\phi_1 = \Diamond_{[0,20]}(F) \wedge \Box \neg (O)\]
then, a possible decomposition $\phi_{1}^{k}$ is given by:
\[\phi_{1}^{1} = \Box (A) \wedge \Diamond_{[0,5]}(A)' \hspace{0.5cm}  [mode: \emph{Take off}]
\] 
\[\phi_{1}^{2} = \Diamond_{[0,5]} (C) \wedge \Box \neg (O) \hspace{0.5cm} [mode: \emph{Steer}]
\]
\[\phi_{1}^{3} = \Diamond_{[0,9]} (F) \wedge \Box \neg (O) \hspace{0.5cm} [mode: \emph{Steer}]
\]
which clearly satisfies Theorem 1. 


\subsection{Final Trajectory Generation}
By decomposing the mission specification $\phi_i$, and by using the MTL to linear constraints translation mechanism, we can replace Problem 1 with a collection of smaller optimization problems, each with a sub-task specification represented as an MTL formula $\phi_i^k$, and an associated linear mode of the hybrid model. 

Here, the linear cost function of choice is $\sum_{t=0}^{T}|u_i(t)|$, where $T$ is the discrete time horizon for the optimal trajectory. Thus, our final formulation of the problem is given by:

\begin{prob}
\[\begin{array}{c c} \underset{x_i,u_i}{\min} \text{ }   &~\sum_{t=0}^{T}|u_i(t)| \\ \text{s.t.} &x_i(t+1)=A_l(t)x_i(t)+B_l(t)u_i(t) \\ &\mathbf{x_i}_{t_0} \models \phi_i^k\\ \end{array}\]
\end{prob}
where $\phi_i^k$ is the MTL specification for the $k^{th}$ sub-task for the $i^{th}$ UAV, $A_l(t)$, $B_l(t)$ are the linear system matrices for the $l^{th}$ mode, and $\mathbf{x_i}_{t_0}$ is the resulting optimal trajectory for the $k^{th}$ sub-task, with $i \in \{1,2,3,...,N\}$, $k \in \{1,2,3,...,M\}$, and $l \in \{1,2,3,...,5\}$. 

For example, in the two UAV setting, for quadrotor $q1$, one sub-task is to go from $A$ to $C$ in 5 time units. The MTL specification for this sub-task is given by $\phi_{sub_{1}^{k}} = \Diamond_{[0,5]} (C) \wedge \Box \neg (O)$, and the associated dynamics are selected from the \emph{Steer} mode. 

Problem 2 represents a collection of MILPs, which can be solved recursively and efficiently using a MILP solver. The resultant trajectories are locally optimal for each individual sub-task, and their existence inherently guarantees safety and finite time completion of the respective sub-tasks. The final trajectory for the complete mission is generated over time by composing all the individual optimal sub-task trajectories. The final path is therefore not optimal but suboptimal with respect to the original mission specification $\phi_i$ for the $i^{th}$ UAV. However, despite this loss of global optimality, the advantages achieved in terms of reduction in computational complexity, and improved scalability with respect to the number of UAVs are far more important, as is shown in the following section.

\section{Simulations and Results}

We apply the proposed method for solving Problem 2 in the same workspace as shown in Fig.~\ref{fig:workspace}. The experiments are run through YALMIP-CPLEX solver using MATLAB interface on an Intel NuC. It is portable computer with an Intel core i7 @ 3.7 GHz CPU, an integrated Intel Iris GPU, and 16 GBs of memory. This setup is directly transferable to a quadrotor as a companion module for onboard computation.
 
We use a $2m$ neighborhood set threshold for the UAVs i.e., $\rho = 2m$. The discrete time horizon for simulation is $T = 30$, and the UAV altitude limit in \emph{Hover} and \emph{Steer} modes is set to $1.5 m$. All dynamics are uniformly discretized at a rate of 5 Hz.

\subsection{Case Study I: Validation (2 UAVs)}

In the two UAV setting (as in \cite{fiaz2019hybrid}), for the mission $\phi_1$, the sub-tasks for the quadrotor $q_1$ are specified as following:
\[\phi_{1}^{1} = \Box (A) \wedge \Diamond_{[0,5]}(A)' \hspace{0.5cm}  [mode: \emph{Take off}]
\] 
\[\phi_{1}^{2} = \Diamond_{[0,5]} (C) \wedge \Box \neg (O) \wedge \Box \neg (q_2) \hspace{0.5cm} [mode: \emph{Steer}]
\]
\[\phi_{1}^{3} = \Diamond_{[0,10]} (F) \wedge \Box \neg (O)  \wedge \Box \neg (q_2)\hspace{0.5cm} [mode: \emph{Steer}]
\]
 \[\phi_{1}^{4} = \Box (F) \wedge \Diamond_{[0,10]} (F)' \hspace{0.5cm} [mode: \emph{Grasp}]
\]
\[\phi_{1}^{5} = \Diamond_{[0,10]}(H_1) \wedge \Box \neg (O)  \wedge \Box \neg (q_2)\hspace{0.5cm} [mode: \emph{Steer}]
\]
\[\phi_{1}^{6} = \Box (H_1) \hspace{0.5cm} [mode: \emph{Land}]
\]
Using the convention defined earlier, the specification $\phi_{1}^{1}$ requires the quadrotor $q_1$ to attain desired threshold altitude (represented as $A'$) while staying inside the 2D region marked $A$. $\phi_{1}^{2}$ requires the quadrotor $q_1$ to reach $C$ within 5 time units, and $\phi_{1}^{3}$ requires it to reach $F$ within 10 time units, while avoiding the obstacle $O$ and the neighboring quadrotor $q_2$. $\phi_{1}^{4}$ requires the UAV to grasp the object at $F$ within 10 time units while staying in $F$, whereas $\phi_{1}^{5}$ asks it to reach $H_1$ within 10 time units. Finally $\phi_{1}^{6}$ forces $q_1$ to stay at $H_1$ indefinitely.

Similarly, the sub-tasks for quadrotor $q_2$ are given as following:
\[\phi_{2}^{1} = \Box (B) \wedge \Diamond_{[0,5]}(B)' \hspace{0.5cm}  [mode: \emph{Take off}]
\]
\[\phi_{2}^{2} = \Diamond_{[0,5]} (C) \wedge \Box \neg (O) \wedge \Box \neg (q_1)\hspace{0.5cm} [mode: \emph{Steer}]
\]
\[\phi_{2}^{3} = \Diamond_{[0,10]} (G) \wedge \Box \neg (O) \wedge \Box \neg (q_1)\hspace{0.5cm} [mode: \emph{Steer}]
\]
 \[\phi_{2}^{4} = \Box (G) \wedge \Diamond_{[0,10]} (G)' \hspace{0.5cm} [mode: \emph{Grasp}]
\]
\[\phi_{2}^{5}= \Diamond_{[0,10]}(H_2) \wedge \Box \neg (O) \wedge \Box \neg (q_1)\hspace{0.5cm} [mode: \emph{Steer}]
\]
\[\phi_{2}^{6} = \Box (H_2) \hspace{0.5cm} [mode: \emph{Land}]
\]
Notice, that the constraint $\Box \neg (q_j)$, where $j \in \cal{N}_i(t)$, $i \in \{1,2\}$, enforces the quadrotors to avoid collision when within $\rho$ proximity of its neighbors. In practice, the UAV that is the first to reach region $C$ and is closest to the window $E$, gets to go through first. The other UAV has to wait in the default \emph{Hover} mode at $C$. 

\begin{figure}
	\centering
		\includegraphics[width=8.8 cm]{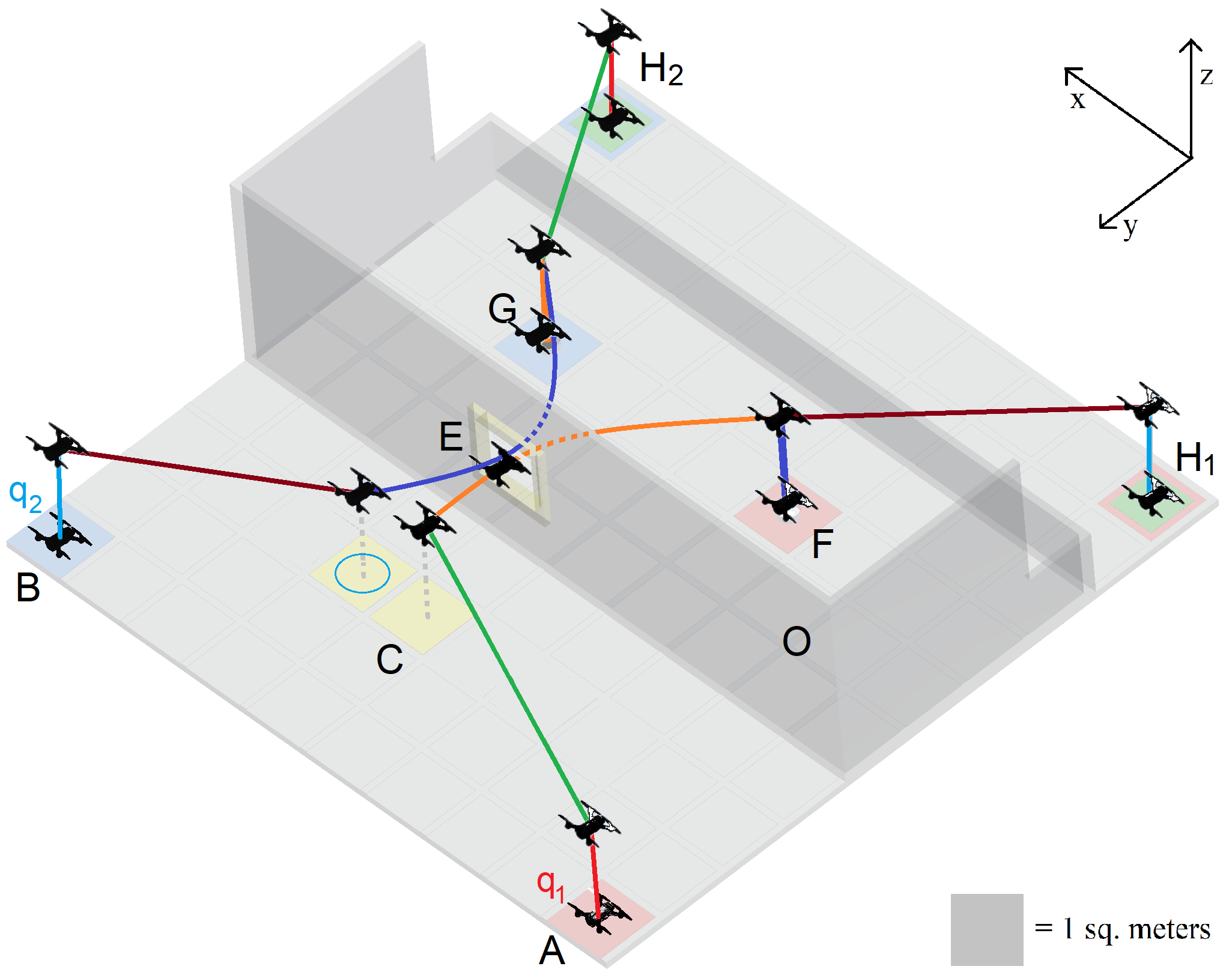}
		\caption{The resultant composed trajectories for the sub-tasks for $q_1$ and $q_2$ operating simultaneously.}
	\label{fig:q1q2}
\end{figure}

Each UAV sequentially solves the set of MILPs for its respective sub-task specifications, and moves along the generated optimal trajectory for the corresponding sub-task. The final mission trajectory is generated by recursive composition of all these optimal sub-task trajectories. Figure~\ref{fig:q1q2} shows the resulting composed trajectories for both the quadrotors operating simultaneously. Both UAVs safely avoid the obstacles and evacuate their respective objects within given finite time limits. As expected, quadrotor $q_2$ waits at $C$. The number of circular rings at $C$ (see Fig.~\ref{fig:q1q2}) correspond to the waiting time for $q_2$ in terms of discrete steps. 

Table~\ref{tab:computation} provides some insight into the timing analysis of each sub-task for both quadrotors. A closer look at this data indicates that all timing constraints are indeed satisfied. Moreover, the computation time for each sub-task indicates that the proposed method can be implemented in real-time. To the best of the authors knowledge, these are one of the fastest computation times reported in the existing MTL based planning literature. The secret to this reduction in computational complexity lies in our divide and conquer approach. From an implementation point of view, the performance can be further improved by using hardware which is optimized for computation (such as Nvidia Jetson TX2 etc.).

\begin{table}[]
\caption{Timing analysis for the $\phi_{i}^{k}$ for $N=2$}
\centering\small
\begin{tabular*}{\linewidth}{@{\extracolsep{\fill}}p{0.275\linewidth}p{0.33\linewidth}p{0.33\linewidth}@{}}
\toprule
$\mathbf{Task~\phi_i^k}$ & \textbf{Computation (sec)} & \textbf{Execution (steps)}\\
\midrule
$\phi_{1}^{1}$ ($A-A'$) & 2.7 &  2 \textcolor{green}{$\leq5$} \\
$\phi_{1}^{2}$ ($A-C$)& 6.3 & 4 \textcolor{green}{$\leq5$} \\
$\phi_{1}^{3}$ ($C-F$)& 10.3 & 7 \textcolor{green}{$\leq10$} \\
$\phi_{1}^{4}$ ($F-F'$)& 3.0 & 3 \textcolor{green}{$\leq10$} \\
$\phi_{1}^{5}$ ($F-H_1$)& 5.7 & 6 \textcolor{green}{$\leq10$} \\
$\phi_{1}^{6}$ ($H_1-H_1'$)& 2.5 & 2 \textcolor{green}{$\leq5$} \\
\midrule
$\phi_{2}^{1}$ ($B-B'$)& 2.7 & 2 \textcolor{green}{$\leq5$} \\
$\phi_{2}^{2}$ ($B-C$)& 5.8 & 4 \textcolor{green}{$\leq5$} \\
$\phi_{2}^{3}$ ($C-G$)& 11.1& 8 \textcolor{green}{$\leq10$} \\
$\phi_{2}^{4}$ ($G-G'$)& 3.0 & 3 \textcolor{green}{$\leq10$} \\
$\phi_{2}^{5}$ ($G-H_2$)& 5.7 & 6 \textcolor{green}{$\leq10$} \\
$\phi_{2}^{6}$ ($H_2-H_2'$)& 2.5 & 2 \textcolor{green}{$\leq5$} \\
\bottomrule
\end{tabular*}
\label{tab:computation}
\end{table}

\subsection{Case Study II: Scalability (N UAVs)}

We now consider the case of $N$ number of UAVs to investigate the scalability features of our method. For this case study, we increase the number of UAVs in the simulation setup, one at a time until one of the finite-time constraints in any of the sub-task specifications is violated for at least one of the UAVs. To keep the analysis consistent, we keep the finite timing constraints for all UAVs the same as before. 

It turns out that for $N=7$, the sub-task specification $\phi_7^3$ fails the satisfaction criterion and hence no solution exists for this sub-task. That is, for $N>6$, one of the UAVs fails to reach its respective target object within the finite time limit of 10 units. The reason is that beyond $N=6$, one access point to the environment is not sufficient to meet the specified timing constraints for all the UAVs, since now the quadrotors have to queue up for longer duration at $C$ in order to avoid collisions at $E$. 

\begin{figure}
	\centering
		\includegraphics[width=8.8 cm]{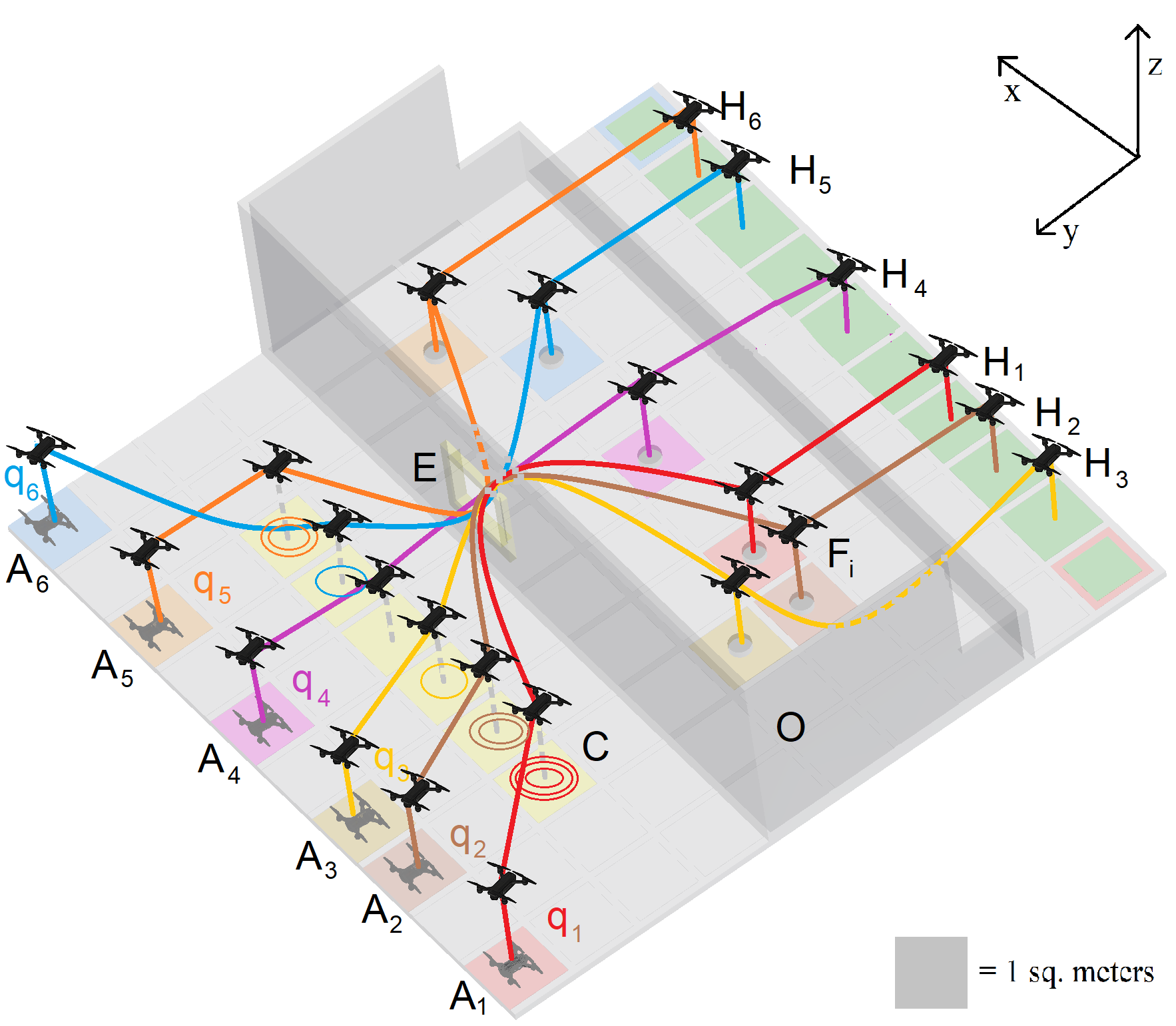}
		\caption{The resultant composed trajectories for the sub-tasks for $N=6$ UAVs operating simultaneously. As before, the number of circular rings at $C$ correspond to the waiting time for the $i^{th}$ quadrotor in terms of discrete steps.}
	\label{fig:qN}
\end{figure}

However, this problem can be solved simply either by increasing the finite time limits for this sub-task, or by executing the mission in an environment with multiple access points. Another way to look at this limitation is to identify that $N=6$ is the maximum number of UAVs that can be deployed successfully under these safety and timing constraints in this particular workspace, and adding more UAVs does not provide any additional value in terms of the success of the mission. Therefore, it is more of a limitation of the workspace than our approach itself. Figure~\ref{fig:qN} shows the resulting composed trajectories for 6 UAVs, in which case all safety and timing constraints are satisfied. Table\ref{tab:computation2} enlists the computation and execution times for the sub-tasks for the $N=6$ case, indicating the satisfaction of all timing constraints.

\begin{table}[]
\caption{Timing analysis for the $\phi_{i}^{k}$ for $N=6$}
\centering\small
\begin{tabular*}{\linewidth}{@{\extracolsep{\fill}}p{0.275\linewidth}p{0.33\linewidth}p{0.33\linewidth}@{}}
\toprule
$\mathbf{Task~\phi_i^k}$ & \textbf{Computation (sec)} & \textbf{Execution (steps)}\\
\midrule
$\phi_{1}^{1}$ ($A_1-A_1'$) & 2.7 &  2 \textcolor{green}{$\leq5$} \\
$\phi_{1}^{2}$ ($A_1-C$)& 5.3 & 4 \textcolor{green}{$\leq5$} \\
$\phi_{1}^{3}$ ($C-F_1$)& 11.3 & 9 \textcolor{green}{$\leq10$} \\
$\phi_{1}^{4}$ ($F_1-F_1'$)& 3.0 & 3 \textcolor{green}{$\leq10$} \\
$\phi_{1}^{5}$ ($F_1-H_1$)& 5.7 & 6 \textcolor{green}{$\leq10$} \\
$\phi_{1}^{6}$ ($H_1-H_1'$)& 2.5 & 2 \textcolor{green}{$\leq5$} \\
\midrule
$\phi_{2}^{1}$ ($A_2-A_2'$)& 2.7 & 2 \textcolor{green}{$\leq5$} \\
$\phi_{2}^{2}$ ($A_2-C$)& 5.8 & 4 \textcolor{green}{$\leq5$} \\
$\phi_{2}^{3}$ ($C-F_2$)& 10.7& 8 \textcolor{green}{$\leq10$} \\
$\phi_{2}^{4}$ ($F_2-F_2'$)& 3.0 & 3 \textcolor{green}{$\leq10$} \\
$\phi_{2}^{5}$ ($F_2-H_2$)& 5.7 & 6 \textcolor{green}{$\leq10$} \\
$\phi_{2}^{6}$ ($H_2-H_2'$)& 2.5 & 2 \textcolor{green}{$\leq5$} \\
\midrule
$\phi_{3}^{1}$ ($A_3-A_3'$)& 2.7 & 2 \textcolor{green}{$\leq5$} \\
$\phi_{3}^{2}$ ($A_3-C$)& 5.8 & 4 \textcolor{green}{$\leq5$} \\
$\phi_{3}^{3}$ ($C-F_3$)& 9.3& 6 \textcolor{green}{$\leq10$} \\
$\phi_{3}^{4}$ ($F_3-F_3'$)& 3.0 & 3 \textcolor{green}{$\leq10$} \\
$\phi_{3}^{5}$ ($F_3-H_3$)& 7.7 & 8 \textcolor{green}{$\leq10$} \\
$\phi_{3}^{6}$ ($H_3-H_3'$)& 2.5 & 2 \textcolor{green}{$\leq5$} \\
\midrule
$\phi_{4}^{1}$ ($A_4-A_4'$)& 2.5 & 2 \textcolor{green}{$\leq5$} \\
$\phi_{4}^{2}$ ($A_4-C$)& 5.8 & 3 \textcolor{green}{$\leq5$} \\
$\phi_{4}^{3}$ ($C-F_4$)& 7.9& 5 \textcolor{green}{$\leq10$} \\
$\phi_{4}^{4}$ ($F_4-F_4'$)& 3.0 & 3 \textcolor{green}{$\leq10$} \\
$\phi_{4}^{5}$ ($F_4-H_4$)& 5.5 & 6 \textcolor{green}{$\leq10$} \\
$\phi_{4}^{6}$ ($H_4-H_4'$)& 2.5 & 2 \textcolor{green}{$\leq5$} \\
\midrule
$\phi_{5}^{1}$ ($A_5-A_5'$)& 2.7 & 2 \textcolor{green}{$\leq5$} \\
$\phi_{5}^{2}$ ($A_5-C$)& 5.8 & 3 \textcolor{green}{$\leq5$} \\
$\phi_{5}^{3}$ ($C-F_5$)& 10.1& 7 \textcolor{green}{$\leq10$} \\
$\phi_{5}^{4}$ ($F_5-F_5'$)& 3.0 & 3 \textcolor{green}{$\leq10$} \\
$\phi_{5}^{5}$ ($F_5-H_5$)& 5.7 & 6 \textcolor{green}{$\leq10$} \\
$\phi_{5}^{6}$ ($H_5-H_5'$)& 2.5 & 2 \textcolor{green}{$\leq5$} \\
\midrule
$\phi_{6}^{1}$ ($A_6-A_6'$)& 2.7 & 2 \textcolor{green}{$\leq5$} \\
$\phi_{6}^{2}$ ($A_6-C$)& 6.3 & 5 \textcolor{green}{$\leq5$} \\
$\phi_{6}^{3}$ ($C-F_6$)& 9.5& 6 \textcolor{green}{$\leq10$} \\
$\phi_{6}^{4}$ ($F_6-F_6'$)& 3.0 & 3 \textcolor{green}{$\leq10$} \\
$\phi_{6}^{5}$ ($F_6-H_6$)& 5.7 & 7 \textcolor{green}{$\leq10$} \\
$\phi_{6}^{6}$ ($H_6-H_6'$)& 2.5 & 2 \textcolor{green}{$\leq5$} \\
\bottomrule
\end{tabular*}
\label{tab:computation2}
\end{table}
\vspace{0.25cm}

\section{Conclusions and Prospects}

We have proposed a hybrid compositional approach to rescue mission planning for quadrotors with MTL task specifications, and have presented an optimization based method which can be implemented in real-time. Using a simple yet realistic search and rescue test case, we have demonstrated the computational efficiency of our approach, and have shown that by breaking down the mission into several sub-tasks, and by using a hybrid model for the system, it is possible to solve the challenging problem of motion planning for multi-agent systems with rich dynamics and finite time constraints in real-time.

In addition to some promising results, this work also poses many new and interesting questions as well. For example, given a finite time constraint for the whole mission, what is the best or optimal way to divide the timing constraints among various sub-tasks. 
Of course it is a scheduling problem, and is dependent on many factors such as robot dynamics, its maximum attainable speed, and nature of the sub-tasks as well. 

In this study, the individual sub-task timing constraints were constructed in a relaxed and uniform fashion. However, it is worth noticing that using a rich dynamical model for the UAV puts less constraints on its maneuverability, and hence can allow it to tackle more conservative finite time constraints as well. For example, the \emph{Steer} mode in our model allows the quadrotor to achieve speeds as high as 1.5 m/s, which is not possible with the usual single mode \emph{Hover} linearization only. 

We anticipate that the scalability features of this approach can be shown with even greater number of UAVs in an environment with multiple access points for instance. Detailed performance comparison of this hybrid approach with some decentralized collision avoidance methods will be beneficial as well. Extension of this work with different tasks, dynamic obstacles other than the UAVs themselves, conservative time constraints, and tolerances in both time and space can also yield interesting results, and are all great directions for future work. 
\vspace{0.1cm}


\bibliography{ifacconf}       
\balance

\end{document}